\documentclass[runningheads]{llncs}
\usepackage{xcolor}
\usepackage{amsmath}
\usepackage{booktabs, multirow}
\usepackage{pifont}
\usepackage[table]{xcolor}
\newcommand{\cmark}{\ding{51}}
\newcommand{\xmark}{\ding{55}}
\usepackage{hyperref}

\usepackage[T1]{fontenc}
\usepackage{graphicx,verbatim}
\usepackage{amssymb}
\begin{document}


\newcommand{\adb}[1]{\textbf{\textcolor{red}{ADB: #1}}}
\newcommand{\rs}[1]{\textbf{\textcolor{blue}{RS: #1}}}
\newcommand{\dr}[1]{\textbf{\textcolor{orange}{DR: #1}}}
\newcommand{\mn}[1]{\textbf{\textcolor{purple}{MN: #1}}}
\newcommand{\ra}[1]{\textbf{\textcolor{gray}{RA: #1}}}
\newcommand{\am}[1]{\textbf{\textcolor{orange}{AM: #1}}}
\newcommand{\ax}[1]{\textbf{\textcolor{pink}{AX: #1}}}

\newcommand{\dataset}{RESPIRE }

%
\title{Stop Holding Your Breath: CT-Informed Gaussian Splatting for Dynamic Bronchoscopy}
\titlerunning{Stop Holding Your Breath}
%

\author{Andrea Dunn Beltran$^{1}$, Daniel Rho$^{1}$, Aarav Mehta$^{1}$, Xinqi Xiong$^{1}$, Raúl San José Estépar$^{2}$, Ron Alterovitz$^{1}$, Marc Niethammer$^{3}$, Roni Sengupta$^{1}$}  
\authorrunning{Anonymized Author et al.}
\institute{$^{1}$University of North Carolina at Chapel Hill 
    \quad $^{2}$Harvard Medical School 
    \quad $^{3}$University of California, San Diego }
  
\maketitle              
\begin{abstract}


Bronchoscopic navigation relies on registering endoscopic video to a preoperative CT scan, but respiratory motion deforms the airway by 5–20 mm, creating CT-to-body divergence that limits localization accuracy. In practice, this is mitigated through breath-hold protocols, which attempt to match the intraoperative anatomy to a static CT, but are difficult to reproduce and disrupt clinical workflow.
We propose to eliminate the need for breath-hold protocols by leveraging patient-specific respiratory modeling.
Paired inhale–exhale CT scans, already acquired for planning, implicitly define the patient-specific deformation space of the breathing airway. By registering these scans, we reduce respiratory motion to a single scalar breathing phase per frame, constraining all reconstructions to anatomically observed configurations.
We embed this representation within a mesh-anchored Gaussian splatting framework, where a lightweight estimator infers breathing phase directly from endoscopic RGB, enabling continuous, deformation-aware reconstruction throughout the respiratory cycle without breath-holds or external sensing.
To enable quantitative evaluation, we introduce RESPIRE, a physically grounded bronchoscopy simulation pipeline with per-frame ground truth for geometry, pose, breathing phase, and deformation. Experiments on RESPIRE show that our approach achieves geometrically faithful reconstruction, over 20× faster training, and 1.22 mm target localization accuracy (within the 3mm clinically relevant tolerances) outperforming unconstrained single-CT baselines.
\textbf{Please check out our website for additional visuals: \url{https://asdunnbe.github.io/RESPIRE/}}

\end{abstract}

\keywords{Bronchoscopy  \and Deformable Reconstruction}

\section{Introduction}
\label{sec:introduction}



Bronchoscopy is a cornerstone of pulmonary medicine, supporting diagnostic biopsy, lesion staging, localised therapy delivery, and airway management.
During a typical procedure, a flexible bronchoscope is advanced through the tracheobronchial tree while the physician navigates using a monocular video stream from the scope's distal tip, guided by a preoperative CT scan.
Accurate navigation is critical for diagnosis of lung cancer, which causes over 120,000 U.S.\ deaths annually; five-year survival exceeds 60\% when diagnosed early but falls below 6\% at stage~IV, yet only 25\% of patients are diagnosed at an early stage~\cite{Siegel2023_Cancer}.
Despite EM tracking and other sensor-based guidance, bronchoscopic navigation remains cognitively demanding and operator-dependent, and a substantial proportion of procedures still fail to reach their intended targets~\cite{Silvestri2020_Evaluation,Yarmus2020_Prospective}.

A fundamental challenge is CT-to-body divergence: the mismatch between the preoperative CT and the intraoperative anatomy, driven primarily by respiratory motion that deforms the airway by 5--20\,mm throughout a procedure~\cite{Pritchett2020_Virtual}. Because navigation relies on registering endoscopic views to CT, this divergence directly causes localization errors and missed targets.
Moreover, camera motion and airway deformation are intrinsically coupled: as the bronchoscope advances, respiratory motion induces coherent global displacement of the airway, making it difficult to disentangle viewpoint changes from underlying anatomical motion.
Breath-hold protocols attempt to freeze the airway in a state that matches the preoperative CT scan, but this is difficult in practice: patients cannot reliably reproduce the breathing state of their preoperative imaging, and repeated breath-holds prolong the procedure and increase patient discomfort.
Existing video-to-CT registration methods~\cite{PANS2024,BREADepth2025,2026breathvl} sidestep this problem by assuming a static airway or localizing within a rigid CT model, leaving the geometric mismatch caused by breathing unresolved.
Deformable reconstruction methods from laparoscopic surgery learn unconstrained per-frame deformation fields, but the bronchoscopic setting differs fundamentally: the camera traverses far greater distances through branching anatomy, compounding the ambiguity between viewpoint change and scene deformation, while respiratory motion displaces the entire visible airway coherently rather than producing isolated tool-tissue perturbations.


We exploit the fact that paired inhale and exhale CT scans, routinely acquired for bronchoscopic planning, already encode the patient-specific deformation space of the breathing airway.
By registering these two scans, we obtain a deformation field and reduce respiratory motion to a single scalar parameter per frame.
We embed this within a mesh-anchored Gaussian splatting framework that recovers the breathing phase directly from endoscopic RGB, enabling continuous deformation-aware video-to-CT registration throughout the breathing cycle without requiring breath-hold protocols or respiratory gating.
Since all CT-based methods already require one scan for initialization, the second CT is a modest additional acquisition that brings endpoint localisation error to 1.22\,mm, within the ${\sim}$3\,mm clinically permissible range for bronchoscopic biopsy targeting, making it possible to maintain accurate registration without breath-holds.

To support evaluation, we introduce \textbf{RESPIRE} a physically grounded simulation pipeline for generating realistic synthetic bronchoscopy videos with per-frame ground truth for depth, pose, breathing phase, and deformation from any pair of inspiration/expiration CTs.
We evaluate against single-CT unconstrained baselines across rendering quality, depth accuracy, breathing phase recovery, and target localization, demonstrating that a second CT resolves the breathing-induced registration error to within clinical tolerance

\section{Related Works} 
\label{sec:related_works}

\paragraph{\textbf{Dynamic Endoscopic Reconstruction.}}
Recent work in endoscopic 3D reconstruction has shifted toward modeling non-rigid tissue deformation. 
Gaussian Splatting–based approaches enable photorealistic dynamic scene reconstruction in minimally invasive surgery, primarily laparoscopic settings~\cite{SurgicalGaussian2024,Deform3DGS2024,Endo4DGS2024,FreeSurGS2024}, with improved robustness through learned deformation priors and temporally consistent updates~\cite{EndoGaussian2025,EndoRDGS2025}. 
These methods, where deformation is driven by instrument teraction, do not generalize to the large camera displacements and global airway deformation encountered in bronchoscopy.
In bronchoscopy, recent approaches leverage CT-derived airway models or monocular depth estimation to enhance geometric understanding~\cite{PANS2024,BREADepth2025}, but typically assume static airway anatomy. 

\paragraph{\textbf{Bronchoscopy Datasets.}}
Compared to laparoscopy, bronchoscopy lacks large-scale datasets with dense geometric measurements. 
Early CT–video registration pipelines aligned virtual bronchoscopy renderings to real video~\cite{Merritt2013CTVideoRegistrationFull}, but did not release depth or deformation ground truth. 
The most established synthetic benchmark, \textit{BronchoPose}~\cite{2023BronchoPose}, provides CT-based airway models with simulated trajectories but assumes static anatomy and uses flat-shaded, uniform-albedo rendering that poorly approximates clinical tissue appearance.
Clinical datasets such as BM-BronchoLC~\cite{Vu2024BMBronchoLCFull} emphasize anatomical landmark and lesion annotation without geometric correspondence, while CT-focused resources like AeroPath~\cite{Stoverud2024AeroPathFull} curate airway segmentations without endoscopic video. 
Overall, current datasets either provide synthetic supervision under static geometry or clinical video without deformation-aware ground truth. 

\section{Methods}
\label{sec:methods}

Given preoperative inspiration and expiration CT scans and a monocular bronchoscopy video with known camera poses, our goal is to reconstruct the airway surface at each frame while jointly estimating the patient's breathing state.
We formulate this as an optimization over a set of mesh-anchored 3D Gaussians whose positions are governed by a single scalar breathing phase $\hat{\alpha}_t \in [0,1]$ that interpolates the airway mesh between two CT-derived anatomical states.

We build upon BridgeSplat~\cite{2025bridgesplat}, a mesh-anchored Gaussian splatting method designed for laparoscopic surgery where a camera observes a confined region and deformation is driven by local tool-tissue interaction.
Bronchoscopy presents a fundamentally different regime: a camera traverses long segments of branching anatomy, continuously encountering new geometry, while respiratory motion displaces the \emph{entire} airway coherently with each breath cycle.
We exploit this by constraining deformation to a one-dimensional subspace defined by preoperative imaging, rather than learning unconstrained per-vertex displacements.

\paragraph{\textbf{Mesh-Anchored Gaussian Representation.}}
\label{sec:method:mesh}
Following~\cite{2025bridgesplat,2024gaussianavatars}, we parameterize 3D Gaussians relative to a CT-derived airway mesh $\mathcal{M} = (\mathbf{V}, \mathbf{F})$.
Each Gaussian $k$ is assigned to a parent triangle $f_k \in \mathbf{F}$ and positioned via learnable barycentric coordinates $\mathbf{b}_k \in \mathbb{R}^3$:
\begin{equation}
    \boldsymbol{\mu}_k = \sum_{j=0}^{2} \lambda_j \, \mathbf{v}_j^{(f_k)}, \quad (\lambda_0, \lambda_1, \lambda_2) = \mathrm{softmax}(\mathbf{b}_k),
    \label{eq:barycentric}
\end{equation}
where $\mathbf{v}_j^{(f_k)}$ are the vertices of face $f_k$ and the softmax constrains each Gaussian to lie within its parent triangle.
Each Gaussian's orientation is aligned with the triangle normal and its normal-direction scale is fixed near zero, producing thin disc primitives that tile the mesh.
The learnable parameters per Gaussian are: barycentric coordinates $\mathbf{b}_k$, major and minor axis length, and first-order spherical harmonic coefficients for view-dependent color; opacity is initialized at creation and frozen throughout optimization.


This anchoring ensures that Gaussian positions, orientations, and covariances are all functions of the mesh vertex positions.
When the mesh deforms, every Gaussian inherits the local stretch, rotation, and translation of its parent triangle automatically, mirroring how tissue itself behaves under respiratory motion, without requiring per-Gaussian deformation parameters.

A differentiable rasterizer $\mathcal{R}$ renders an image $\hat{\mathbf{I}}_t$ from Gaussians $\mathcal{G}$, the deformed mesh $\mathbf{V}(\hat{\alpha}_t)$, and camera pose $\mathbf{T}_t$.
The photometric loss backpropagates through $\mathcal{R}$,
enabling joint optimization of appearance and respiratory state $\hat{\alpha}_t$.


\paragraph{\textbf{CT-Informed Breathing Deformation.}}
\label{sec:method:deformation}
Given registered inspiration and expiration meshes $\mathbf{V}_{\text{insp}}, \mathbf{V}_{\text{exp}} \in \mathbb{R}^{N_v \times 3}$ with identical topology obtained via deformable CT-to-CT registration, we precompute a fixed displacement field $\boldsymbol{\Delta} = \mathbf{V}_{\text{exp}} - \mathbf{V}_{\text{insp}}$.
The mesh at any breathing phase $\hat{\alpha}_t \in [0,1]$ is:
\begin{equation}
    \mathbf{V}(\hat{\alpha}) = \mathbf{V}_{\text{insp}} + \hat{\alpha}_t \cdot \boldsymbol{\Delta}, \quad \hat{\alpha}_t \in [0,1] =\text{ [full inspiration, full expiration]},
    \label{eq:breathing}
\end{equation}
Because $\boldsymbol{\Delta}$ is derived from the patient's own imaging, every predicted configuration lies on the anatomically observed trajectory between two real respiratory states.
This eliminates the need for the geometric regularization (ARAP, isometric, rigidity, and visibility losses) required by unconstrained approaches~\cite{2025bridgesplat} to prevent implausible deformations.

\paragraph{Activation function.}
Having defined how the mesh deforms given a breathing phase (Eq.~\ref{eq:breathing}), we now describe how $\hat{\alpha}_t$ is parameterized for optimization.
Estimating the breathing phase $\hat{\alpha}_t$ for each frame $t$ requires a bounded parameterization.
During optimization, the photometric loss backpropagates through the rasterizer, mesh deformation, and Gaussian positions to the phase parameter. As a result, the parameterization must provide stable gradients across the entire breathing cycle, including at the extrema where accurate alignment is most critical.
A standard sigmoid mapping enforces the desired bounds but suffers from saturation near $0$ and $1$, leading to vanishing gradients precisely at full inspiration and expiration. This makes it difficult to correct phase errors at these clinically important states.
Instead, we instead adopt a cosine-based parameterization with a linear gradient leak:
%
\begin{equation}
    \hat{\alpha}_t = (1 - \varepsilon)\,\frac{1}{2}\!\left(1 - \cos\theta_t\right) + \varepsilon\,\frac{\theta_t}{\pi}, \quad \theta_t \in [0, \pi],
    \label{eq:cosine_activation}
\end{equation}
where $\varepsilon{=}0.05$.
The cosine term provides a smooth, monotonic traversal of $[0,1]$ consistent with the periodic nature of breathing, while the linear component ensures a non-zero gradient of at least $\varepsilon/\pi$ throughout the domain. This prevents optimization from stalling at the breathing extremes, analogous to how leaky activations mitigate dead neurons in deep networks.


\paragraph{\textbf{Breathing Phase Estimation.}}
\label{sec:method:phase}
For the first frame, where no appearance model is yet available, $\hat{\alpha}$ is initialized via a depth-based grid search to ensure a geometrically consistent starting point; all subsequent frames estimate $\hat{\alpha}_t$ from RGB supervision alone, reflecting clinical reality where ground-truth depth is rarely available.
Per-frame optimization follows a three-phase schedule: (i)~optimize only $\theta_t$ with appearance frozen, (ii)~optimize only appearance with $\theta_t$ frozen, and (iii)~joint optimization.
Updating $\hat{\alpha}$ deforms the mesh via Eq.~\eqref{eq:breathing}, which repositions all Gaussian centres through their barycentric anchoring; updating appearance refines the per-Gaussian tangential scales, barycentric coordinates, and spherical harmonic color coefficients at fixed geometry.
The phased strategy prevents the appearance parameters from absorbing geometric errors that should be explained by the breathing phase.


\paragraph{\textbf{Loss Function.}}
\label{sec:method:loss}
The optimization objective combines photometric fidelity and temporal regularization:
\begin{equation}
    \mathcal{L} = \underbrace{w_c \left[(1 - w_s)\,\|\hat{\mathbf{I}} - \mathbf{I}\|_1 + w_s\,(1 - \text{SSIM}(\hat{\mathbf{I}}, \mathbf{I}))\right]}_{\mathcal{L}_{\text{photo}}} + \underbrace{w_t\,(\hat{\alpha}_t - \hat{\alpha}_{t-1})^2}_{\mathcal{L}_{\text{temp}}},
    \label{eq:loss}
\end{equation}
%
where $w_c$, $w_s$, $w_t$ are loss weights. 
This two-term objective is a direct consequence of our formulation: because the deformation field is anatomically grounded, the multiple additional geometric regularization terms required by unconstrained approaches are unnecessary, reducing both implementation complexity and computational cost.

\begin{figure}
\vspace{-0.75em}

    \centering
    \includegraphics[width=0.9\linewidth]{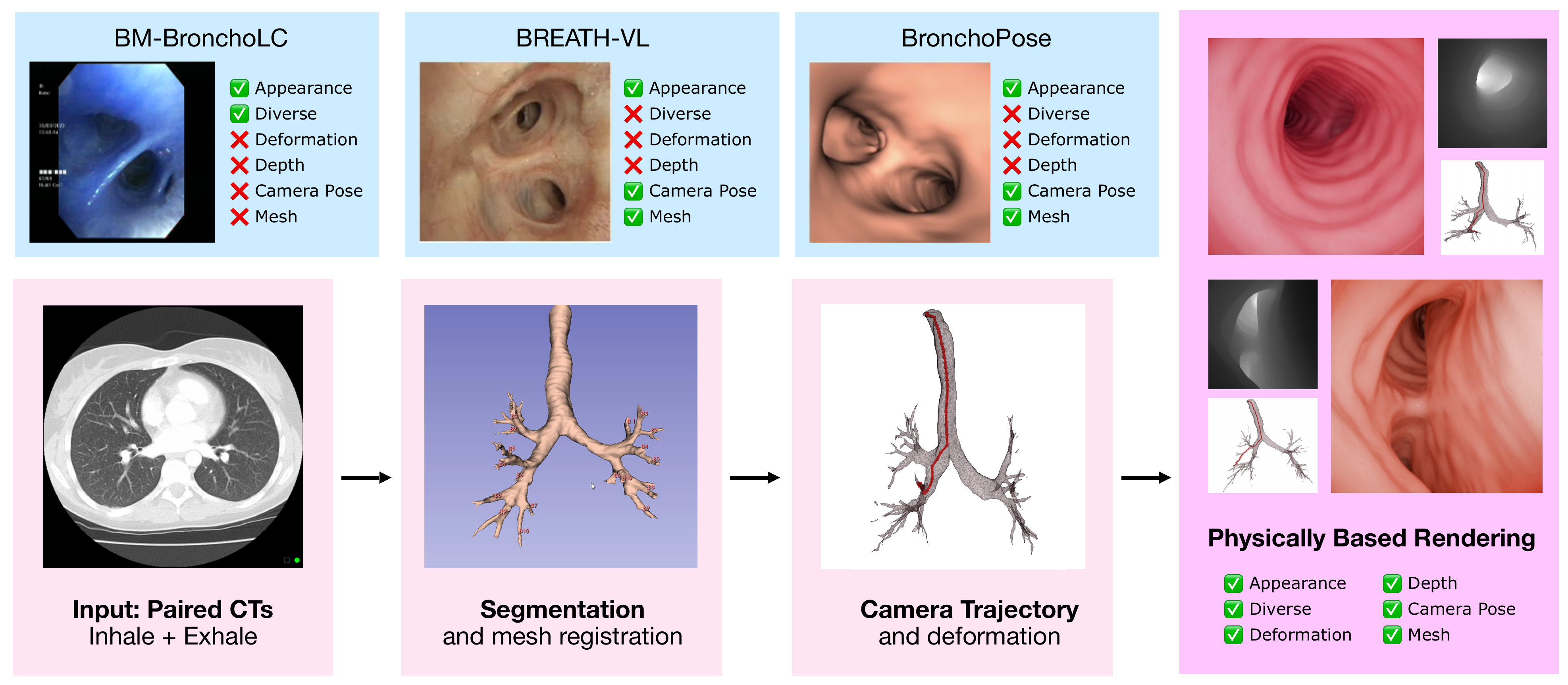}
    \caption{\textbf{RESPIRE Overview.} \textit{Top:} Existing bronchoscopy datasets lack dense geometric and respiratory ground truth; RESPIRE provides all six annotation types. \textit{Bottom:} The RESPIRE pipeline generates realistic bronchoscopy sequences with per-frame annotations from paired inspiration/expiration CTs.}
    \label{fig:data}
    \vspace{-0.5em}

\end{figure}

\vspace{-0.25em}

\section{RESPIRE Framework}
\label{sec:dataset}

\noindent Existing bronchoscopy datasets do not support evaluation of deformation-aware reconstruction. Synthetic benchmarks assume static airway geometry, while clinical datasets lack dense geometric correspondence, respiratory state, and ground-truth deformation. As a result, it is not possible to quantitatively assess whether a method correctly models breathing-induced motion.

To address this, we introduce \textbf{\dataset} (\textbf{R}espiratory \textbf{E}ndoscopic \textbf{S}cenes with \textbf{P}aired \textbf{I}nspiration-expiration CTs for \textbf{RE}construction), a physically grounded simulation framework that generates realistic bronchoscopic video with per-frame respiratory deformation from any pair of inspiration/expiration CT scans.
The pipeline takes segmented airway meshes as input and produces RGB frames, dense depth maps, 6-DoF camera poses, intrinsics, breathing phase, and intermediate deformed meshes.
Because the framework is agnostic to the source of the paired CTs, it scales naturally as more data becomes available across institutions and imaging protocols. \dataset enables direct evaluation of deformation-aware reconstruction under respiratory motion, which is not possible with existing datasets.





\paragraph{\textbf{Mesh Registration and Breathing Model.}}
\label{sec:dataset:mesh}
Each case is derived from paired thoracic CT scans acquired at full inspiration and end expiration within the same imaging session to ensure a shared canonical space. Lung volumes are segmented from both scans to extract airway meshes; segmentation quality directly governs deformation field fidelity.
To establish dense correspondence between inspiration and expiration meshes, we perform deformable registration using GradICON~\cite{2023gradicon}. 
Intermediate breathing states are obtained by interpolating:
%
\begin{equation}
    \mathbf{V}(\alpha) = (1-\alpha)\,\mathbf{V}_{\text{insp}} + \alpha\,\mathbf{V}_{\text{exp}}, \quad \alpha \in [0,1].
    \label{eq:mesh_interp}
\end{equation}
%
This ensures that all simulated motion lies within anatomically observed patient-specific deformation, rather than synthetic or unconstrained perturbations.

This interpolation is consistent with standard 4D-CT respiratory motion
modelling~\cite{castillo2009framework,werner2009patient} and provides a
stable first-order approximation across the navigable airway tree.
The respiratory phase $\alpha(t)$ follows a periodic profile modeled on tidal breathing pressure--volume curves~\cite{West2012RespPhysEssentials}, with a default 1.5\,s inhalation and 2.5\,s exhalation.
Patient-specific temporal scaling introduces inter-subject variability in breathing rate.

\paragraph{\textbf{Camera Trajectory.}}
\label{sec:dataset:trajectory}
Realistic camera motion is critical, as appearance changes induced by viewpoint can be confounded with deformation. 
Camera trajectories follow Voronoi-based airway centerlines extracted via VMTK~\cite{Izzo2018VMTK}.
For each trajectory, a distal bronchiolar endpoint is randomly selected and the shortest centerline path to the trachea is computed.
The camera advances at a configurable speed (default 10\,mm/s)~\cite{Merritt2013CTVideoRegistrationFull} with the optical axis aligned to the local tangent.
Endpoint selection, speed, and trajectory count are user-adjustable to match specific procedural scenarios.
At each frame, the mesh is deformed to the current $\alpha(t)$ via Eq.~\eqref{eq:mesh_interp} prior to rendering.

\paragraph{\textbf{Physically Based Rendering}}
\label{sec:dataset:rendering}


    
    


Photorealistic rendering is essential, as breathing phase is inferred directly from RGB appearance; unrealistic shading or texture would bias phase estimation and limit transfer to real data. Prior synthetic bronchoscopy data relies on flat shading and uniform albedo, producing images far removed from clinical footage~\cite{2023BronchoPose}.

We render in Blender~\cite{BlenderSoftware} using the Cycles path-tracing engine with a custom tissue material incorporating \textbf{subsurface scattering} to reproduce mucosal translucency, \textbf{specular reflections} to capture the wet-surface glare characteristic of bronchoscope illumination, and \textbf{spatially varying albedo} to model patient-specific vascular patterning. A point light co-located with the camera replicates the radial intensity falloff of fibre-optic bronchoscope optics~\cite{Yarmus2010Bronchoscopes}. To further approximate clinical imaging, we introduce frame-to-frame variation in light intensity to mimic automatic exposure and white balance adjustments in bronchoscopic systems. Without this, near-wall views at close proximity produce saturated, washed-out regions, which are not representative of real acquisitions.

\vspace{1em}
\noindent Together, these components provide the first framework that jointly models realistic appearance, respiratory deformation, and ground-truth geometry, enabling quantitative evaluation of deformation-aware bronchoscopic reconstruction.

\section{Experiments}
\label{sec:experiment}

\paragraph{\textbf{Dataset.}}
We evaluate on \textbf{nine cases} from the COPDGene dataset~\cite{COPD} spanning varied airway morphologies and deformation magnitudes. Bronchial trees were segmented using a semi-supervised nnU-Net from paired inspiration/expiration CTs, and the \dataset\ pipeline (Sec.~\ref{sec:dataset}) generates over 400 frames per case with varying breathing phase, tissue materials, and camera trajectories. This setup enables controlled evaluation of deformation-aware reconstruction under realistic respiratory motion, with dense ground truth for geometry, pose, and breathing phase that is not available in clinical data.

\paragraph{\textbf{Metrics.}}
We evaluate along five complementary axes, each targeting a distinct aspect of the reconstruction problem.
\begin{itemize}
    \item \textit{Rendering quality} (PSNR, SSIM) measures photometric fidelity of the reconstructed scene.  
    \item \textit{Depth reconstruction} (RMSE, $\delta{<}1.25$) evaluates local geometric accuracy against ground-truth depth.  
    \item \textit{Breathing phase estimation} (MAE, Pearson $r$) assesses the accuracy of recovered respiratory dynamics, enabled by \dataset.  
    \item \textit{Target localization accuracy} measures clinically relevant endpoint precision for biopsy targeting. We measure how accurately we can predict the distance between the endoscopy and the target in addition to the reconstruction accuracy of the airway tree. To achieve this, we slice both the ground-truth and predicted meshes at the camera's image plane at the final frame and report the RMSE between the two contours. Additionally, we measure the 3D displacement of a single contour vertex, simulating the localization error when directing a biopsy instrument toward a target on the airway wall. 
    \item \textit{Runtime} (total time, per-frame time) captures computational efficiency.
\end{itemize}

Because baseline methods do not model respiratory state, breathing and target metrics are only applicable to methods with explicit deformation modeling or mesh representations.

\paragraph{\textbf{Baselines and Ablations.}}
We compare four settings varying along two axes: single vs.\ paired-CT deformation, and mesh-anchored vs.\ free Gaussians.

\textbf{BridgeSplat}~\cite{2025bridgesplat} learns unconstrained per-vertex deformation regularized by ARAP, isometric, and rigidity losses.  
\textbf{BridgeSplat w/o mesh} removes mesh anchoring.  
\textbf{Ours w/o mesh} applies our paired-CT breathing model to free Gaussians.  
\textbf{Ours} combines paired-CT deformation with mesh anchoring.

This design isolates the contributions of (i) constraining deformation to a CT-derived subspace and (ii) enforcing geometric consistency through mesh anchoring.

\subsection{Results}

\noindent The table below summarizes results across nine cases. Our method consistently outperforms both baselines across rendering, depth reconstruction, and runtime, while uniquely recovering breathing phase.

\begin{table*}                                                
\centering           
\caption{Quantitative comparison on \dataset\ (mean over 9 cases). Best in \textbf{bold}, $\uparrow$\,=\,higher is better, $\downarrow$\,=\,lower is better. Breathing and target metrics apply only to methods that model respiratory state or maintain an explicit mesh, respectively.}      
\label{tab:main_results}   
\resizebox{\textwidth}{!}{%
\begin{tabular}{@{}lcc|cc|cc|cc|cc|cc@{}}         
\toprule                            
& \multicolumn{2}{c|}{\textbf{Setting}}   
& \multicolumn{2}{c|}{\textbf{Rendering}} 
& \multicolumn{2}{c|}{\textbf{Depth}}  
& \multicolumn{2}{c|}{\textbf{Breathing}}
& \multicolumn{2}{c|}{\textbf{Target}}  
& \multicolumn{2}{c}{\textbf{Runtime}} \\   
\cmidrule(lr){2-3} \cmidrule(lr){4-5} \cmidrule(lr){6-7} \cmidrule(lr){8-9} \cmidrule(lr){10-11} \cmidrule(lr){12-13}                       
\textbf{Method}  
& CTs & Mesh       
& PSNR$\uparrow$ & SSIM$\uparrow$  
& RMSE$\downarrow$ & $\delta{<}1.25$$\uparrow$               
& MAE$\downarrow$ & Pearson $r$$\uparrow$
& Contour$\downarrow$ & Target$\downarrow$ 
& Time$\downarrow$ & s/frame$\downarrow$ \\
\midrule                                                                                   
BridgeSplat~\cite{2025bridgesplat} & 1 & \cmark & 20.65 & 0.683 & 53.8 & 0.56 & -     & -     & 5.67 & 5.61 & 383.4\,min  & 44.24 \\
BridgeSplat~\cite{2025bridgesplat} & 1 & \xmark & 27.80 & 0.846 & 26.9 & 0.50 & -     & -     & -     & -     & 96.8\,min   & 11.26 \\                                                                         
\midrule                 
Ours (no mesh)                     & 2 & \xmark & \textbf{30.71} & \textbf{0.924} & 5.9 & 0.79 & 0.162 & 0.811 & -     & -     & \textbf{13.4\,min}  & \textbf{1.73} \\                                                                      
\rowcolor[gray]{0.92}
Ours                               & 2 & \cmark & 27.37 & 0.821 & \textbf{3.2} & \textbf{0.88} & \textbf{0.158} & \textbf{0.833} & \textbf{1.24} & \textbf{1.22} & 17.4\,min & 1.47 \\                                                                  
\bottomrule
\end{tabular}%
}                                                            
\end{table*}

\begin{figure}
\vspace{-0.75em}
    \centering
    \includegraphics[width=0.90\linewidth]{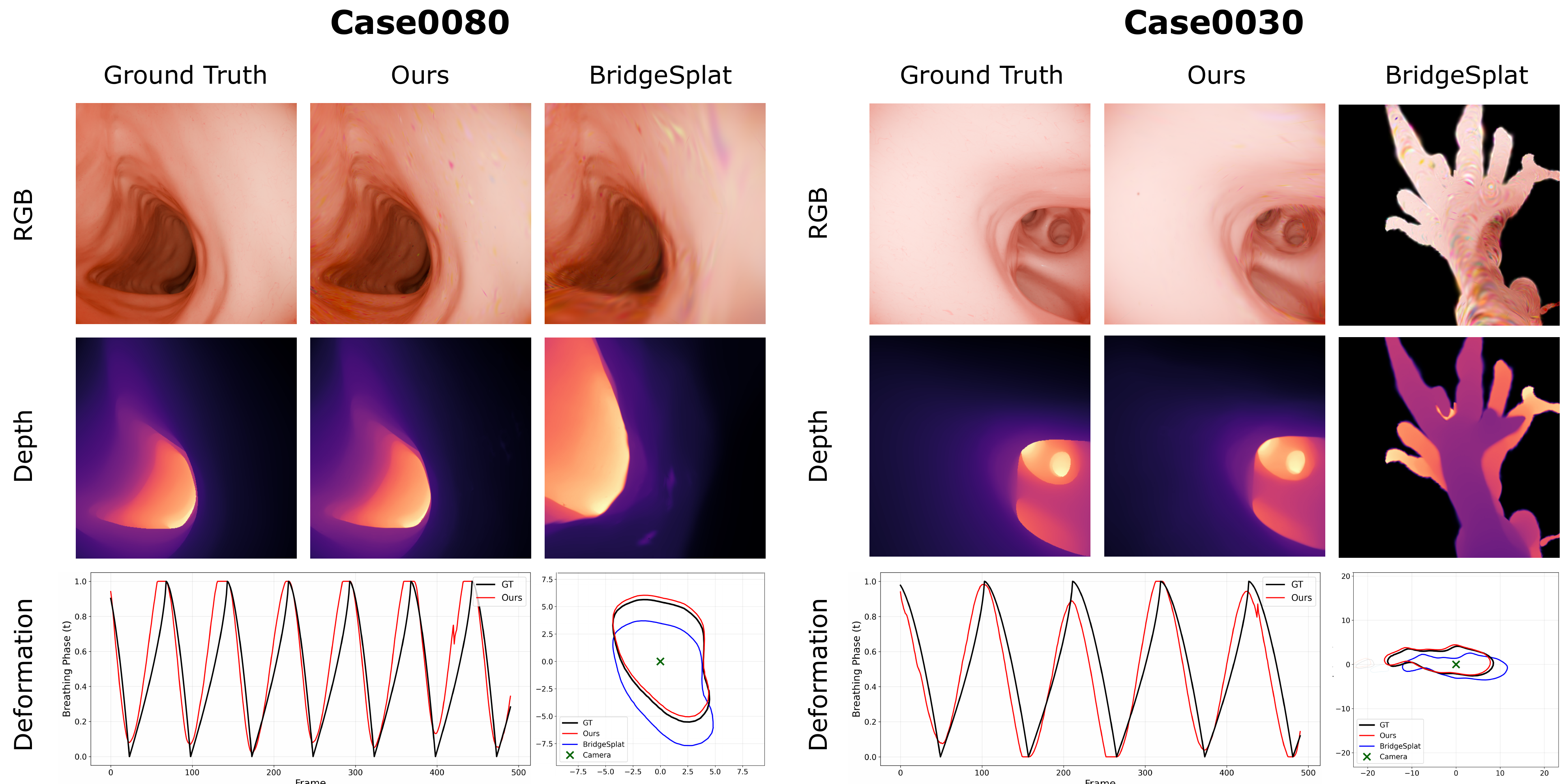}
    \caption{\textbf{RESPIRE Results.} 
    \textit{Top:} rendered RGB and depth versus ground truth. Our method maintains geometric fidelity under respiratory deformation; BridgeSplat produces plausible appearance but incorrect geometry, with the mesh drifting outside the airway lumen in distal segments. \textit{Bottom:} recovered breathing phase trajectory and endpoint contour overlay, showing sub-millimeter alignment at the final frame.}
    \label{fig:results}
    \vspace{-01.0em}
\end{figure}

\paragraph{\textbf{Rendering and depth.}}
Our method reduces depth RMSE by 94\% over BridgeSplat, with 32\% and 20\% improvements in PSNR and SSIM. These gains arise primarily from constraining deformation to a patient-specific anatomical trajectory, which eliminates the ambiguity between appearance and geometry present in unconstrained methods.
Mesh anchoring provides an additional 46\% reduction in RMSE by preventing Gaussians from satisfying the photometric loss through off-surface drift, instead enforcing physically consistent deformation (Fig.~\ref{fig:results}).

\paragraph{\textbf{Breathing phase.}}
Our method recovers the respiratory cycle with MAE\,=\,0.158 and Pearson $r$\,=\,0.833 from RGB supervision alone. To our knowledge, this is the first quantitative evaluation of breathing phase estimation in bronchoscopy.
By explicitly modeling respiratory dynamics, the reconstruction adapts continuously to the current anatomical state, rather than relying on breath-hold protocols to approximate a static CT.

\paragraph{\textbf{Target localization.}}
The clinical goal of navigational bronchoscopy is accurate targeting for biopsy. We evaluate this by slicing both predicted and ground-truth meshes at the final camera plane and computing contour RMSE, as well as the 3D displacement of a representative target point.
Our method achieves a contour RMSE of 1.24\,mm and target error of 1.22\,mm, well within the ${\sim}$3\,mm precision required for bronchoscopic biopsy targeting. At this level of accuracy, breathing-induced CT-to-body divergence is no longer the limiting factor in navigation, effectively eliminating the need for breath-hold protocols during tissue sampling.

\paragraph{\textbf{Runtime.}}
Our method reduces training time compared to BridgeSplat (17.4\,min vs.\ 383.4\,min, 22$\times$ faster) by replacing per-frame unconstrained deformation with a single scalar breathing parameter over a precomputed deformation field.
\section{Conclusion}
\label{sec:conclusion}


We have shown that respiratory deformation in bronchoscopy can be modeled as a low-dimensional, patient-specific process grounded in paired CT scans that are already acquired in clinical workflows. By reducing motion to a single scalar breathing phase and constraining reconstruction to anatomically observed configurations, our approach improves geometric accuracy, rendering fidelity, and computational efficiency, while recovering a clinically interpretable respiratory trajectory directly from RGB video.
Beyond methodological improvements, this formulation addresses a fundamental ambiguity in bronchoscopic reconstruction: the coupling between camera motion and airway deformation. By constraining motion to a physiologically valid subspace, we enable stable, deformation-aware reconstruction without reliance on breath-hold protocols. The RESPIRE framework further makes this problem measurable for the first time, providing dense ground truth for geometry, pose, and respiratory dynamics.

\paragraph{\textbf{\textit{Limitations and Future Work.}}}
RESPIRE depends on accurate airway segmentation, which may require manual refinement in distal or pathological regions. While our physically based rendering improves realism, a domain gap to clinical video remains and may impact generalization. The linear deformation model captures first-order respiratory motion but does not represent nonlinear pressure--volume dynamics at intermediate phases~\cite{werner2009patient}. Finally, we assume known camera pose; jointly estimating pose and breathing phase from monocular video remains an important direction toward full clinical deployment.
Future work will focus on reducing the simulation-to-clinical domain gap, incorporating higher-order or learned deformation models, and integrating pose estimation into the reconstruction pipeline to enable fully self-supervised, deformation-aware bronchoscopic navigation.

\section{Acknowledgements}

This work is supported by a National Institute of Health (NIH) project \#R21EB035832 "Next-gen 3D Modeling of Endoscopy Videos" and \#R21EB037440 "Gen-AI Airway Simulator for 3D Endoscopy"



%
%
%
\clearpage
\bibliographystyle{splncs04}
\bibliography{bib,UNCRobotics}

\end{document}